\documentclass{article}

\PassOptionsToPackage{authoryear}{natbib}

 \usepackage[final]{nips_2016}
\usepackage[utf8]{inputenc} 
\usepackage[T1]{fontenc}    
\usepackage{hyperref}       
\usepackage{url}            
\usepackage{booktabs}       
\usepackage{nicefrac}       
\usepackage{microtype}      

\usepackage{xfrac}
\usepackage{sidecap}
\usepackage{subfigure}
\usepackage{scrextend} 
\usepackage{amsmath}
\usepackage{bm}
\usepackage{array}
\newcolumntype{P}[1]{>{\centering\arraybackslash}p{#1}}

\DeclareMathOperator{\softplus}{softplus}
\DeclareMathOperator{\softmax}{softmax}
\DeclareMathOperator{\sig}{sigmoid}
\DeclareMathOperator{\taylor}{taylor\_softmax}

\title{The Z-loss: a shift and scale invariant classification\\ loss belonging to the Spherical Family}

\author{
  Alexandre de Brébisson \\
  MILA, University of Montréal\\
  \texttt{alexandre.de.brebisson@umontreal.ca} \\
\And
Pascal Vincent \\
MILA, University of Montréal and CIFAR\\
\texttt{vincentp@iro.umontreal.ca} \\
}

\begin{document}

\maketitle

\begin{abstract}

Despite being the standard loss function to train multi-class neural networks, the log-softmax has two potential limitations. First, it involves computations that scale linearly with the number of output classes, which can restrict the size of problems that we are able to tackle with current hardware. Second, it remains unclear how close it matches the \emph{task loss} such as the top-k error rate or other non-differentiable evaluation metrics which we aim to optimize ultimately. In this paper, we introduce an alternative classification loss function, the Z-loss, which is designed to address these two issues. Unlike the log-softmax, it has the desirable property of belonging to the \emph{spherical loss family} \citep{vincent2015efficient}, a class of loss functions for which training can be performed very efficiently with a complexity independent of the number of output classes. We show experimentally that it significantly outperforms the other \emph{spherical} loss functions previously published and investigated. Furthermore, we show on a word language modeling task that it also outperforms the log-softmax with respect to certain ranking scores, such as top-k scores, suggesting that the Z-loss has the flexibility to better match the \emph{task loss}. These qualities thus makes the Z-loss an appealing candidate to train very efficiently large output networks such as word-language models or other extreme classification problems. On the One Billion Word \citep{ChelbaMSGBKR14} dataset, we are able to train a model with the Z-loss 40 times faster than the log-softmax and more than 4 times faster than the  hierarchical softmax.

\end{abstract} 

\vspace{-1em}
\section*{Introduction}
\vspace{-0.5em}

Classification tasks are usually associated to a loss function of interest, the \emph{task loss}, which we aim to minimize ultimately. \emph{Task losses}, such as the classification error rate, are most of the time non-differentiable, in which case a differentiable \emph{surrogate loss} has to be designed so that it can be minimized with gradient-descent. This \emph{surrogate loss} act as a proxy for the \emph{task loss}: by minimizing it, we hope to minimize the \emph{task loss}. 

The most common \emph{surrogate loss} for multi-class classification is the negative log-softmax, which corresponds to maximizing the log-likelihood of a probabilistic classifier that computes class probabilities with a softmax. Despite being ubiquitous, it remains unclear to which degree it matches the \emph{task loss} and why the softmax is being used rather than alternative normalizing functions. Traditionally, other loss functions have also been used to train neural networks for classification, such as the mean square error after sigmoid with 0-1 targets, or the cross-entropy after sigmoid, which corresponds to each output being modeled independently as a Bernoulli variable. Multi-class generalisation of margin losses \citep{lapin2015svm} and ranking losses \citep{usunier2009ranking, Weston-ijcai2011} can also be used when a probabilistic interpretation is not required.
 Although these loss functions appear to perform similarly on small scale problems, they seem to behave very differently on larger output problems, such as neural language models~\citep{nnlm:2001:nips}. Therefore, in order to better evaluate the difference between the loss functions, we decided to focus our experiments on language models with a large number of output classes (up to 793471). Note that computations for all these loss functions scale linearly in the number of output classes. 

In this paper, we introduce a new loss function, the Z-loss, which, contrary to the log-softmax or other mentioned alternatives, has the desirable property of belonging to the \emph{spherical family} of loss functions, for which the algorithmic approach of~\cite{vincent2015efficient} allows to compute the exact gradient updates in time and memory complexity independent of the number of classes. If we denote $d$ the dimension of the last hidden layer and $D$ the number of output classes, for a spherical loss, the exact updates of the output weights can be computed in $O(d^2)$ instead of the naive $O(d \times D)$ implementation, i.e. independently from the number of output classes $D$. The gist of the algorithm is to replace the costly dense update of output matrix $W$ by a sparse update of its factored representation $VU$ and 
to maintain summary statistics of $W$ that allow computing the loss in $O(d^2)$. We refer the reader to the aforementioned paper for the detailed description of the approach. Several spherical loss functions have already been investigated \citep{brebisson2016spherical} but they do not seem to perform as well as the log-softmax on large output problems.


Several other workarounds have been proposed to tackle the computational cost of huge softmax layers and can be divided in two main approaches. The first are sampling-based approximations, which compute only a tiny fraction of the output's dimensions \citep{Gutmann+Hyvarinen-2010, Mikolov-et-al-NIPS2013, Mnih2013, NIPS2014_5329, ji2015blackout}. The second is the hierarchical softmax, which modifies the original architecture by replacing the large output softmax by a heuristically defined hierarchical tree \citep{Morin+al-2005, Mikolov-et-al-NIPS2013}. \cite{chen2015strategies} benchmarked many of these methods on a language modeling task and among those they tried, they found that for very large vocabularies, the hierarchical softmax is the fastest and the best for a fixed budget of training time. Therefore we will also compare the Z-loss to the hierarchical softmax.


\textbf{Notations:} In the rest of the paper, we consider a neural network with $D$ outputs. We denote by $\bm{o} = [o_1,...,o_k,...,o_D]$ the output pre-activations, i.e. the result $\bm{o} = W \bm{h}$ of the last matrix multiplication of the network, where $\bm{h}$ is the representation of the last hidden layer. $c$ represents the index of the target class, whose corresponding output activation is thus $o_c$.

\vspace{-0.5em}
\section{Common multi-class neural network loss functions}
\vspace{-0.25em}

In this section, we briefly describe the different loss functions against which we compare the Z-loss.

\vspace{-1em}
\subsection{The log-softmax loss function}
\vspace{-0.5em}

The standard loss function for multi-class classification is the log-softmax, which corresponds to minimizing the negative log-likelihood of a softmax model. The $\softmax$ activation function models the output of the network as a categorical distribution, its $i^{th}$ component being defined as $
\softmax_i(\bm{o}) = \sfrac{\exp(o_i)} {\sum_{k=1}^D \exp(o_k)}$. We note that the softmax is invariant to shifting $\bm{o}$ by a constant but not to scaling. Maximizing the log-likelihood of this model corresponds to minimizing the classic log-softmax loss function $L_S$:
\begin{align*}
L_S(\bm{o}, c) &= - \log \softmax_c(\bm{o}) = -o_c + \log \sum_{k=1}^D \exp(o_k),
\end{align*}
whose gradient is $\frac{\partial L_S}{\partial o_{c}} = -1 + \softmax_c(\bm{o})$ and $ \frac{\partial L_S}{\partial o_{k}}_{k \neq c} = \softmax_k(\bm{o})$. Intuitively, mimimizing this loss corresponds to maximizing $o_c$ and minimizing the other $o_k$. Note that the sum of the gradient components is zero, reflecting the competition between the activations $o$.

%

\vspace{-1em}
\subsection{Previously investigated spherical loss functions}
\label{spherical_losses}
\vspace{-0.5em}

Recently, \cite{vincent2015efficient} proposed a novel algorithmic approach to compute the exact updates of the output weights in a very efficient fashion, independently of the number of classes, provided that the loss belongs to a particular class of functions, called the \emph{spherical family}. This family is composed of the functions that can be expressed using only $o_{c}$,  the squared
norm of the whole output vector $\sum_{i}^D o_{i}^{2}$ and $\sum_{i}^D o_{i}$:
\begin{align*}
\mathcal{L}\left( \sum_{i}^D o_{i}, \sum_{i}^D o_{i}^{2}, o_{c} \right).
\end{align*}

\textbf{The Mean Square Error}: The MSE after a linear mapping (with no final sigmoid non-linearity) is the simplest member of the spherical family. It is defined as $L_{MSE}(\bm{o}, c) = \frac{1}{2} \sum_{k=1}^D (o_k - \delta_{kc})^2$. The form of its gradient is similar to the log-softmax and its components also sums to zero: $\frac{\partial L_{MSE}}{\partial o_{c}} = -1 + o_c$ and $
\frac{\partial L_{MSE}}{\partial o_{k}}_{k \neq c} = -o_k$. Contrary to the softmax, the MSE penalizes overconfident high values of $o_c$, which is known to slow down training.

\textbf{The log-Taylor-softmax}: Several loss functions belonging to the spherical family have recently been investigated by  \cite{brebisson2016spherical}, among which the Taylor Softmax was retained as the best candidate. It is obtained by replacing the exponentials of the softmax by their second-order Taylor expansions around zero:
\[
\taylor_i(\bm{o}) = \frac{1 + o_i + \frac{1}{2} o_i^2}{\sum_{k=1}^n (1 + o_k + \frac{1}{2} o_k^2)}. 
\]
The components are still positive and sum to one so that it can model a categorical distribution and can be trained with maximum likelihood. We will refer to this corresponding loss as the Taylor-softmax loss function:
\[
L_T(\bm{o}, c) = - \log(\taylor_c(\bm{o})). 
\]
Although the Taylor softmax performs slightly better than the softmax on small output problems such as MNIST and CIFAR10, it does not scale well with the number of output classes \citep{brebisson2016spherical}.

\vspace{-1em}
\subsection{Hierarchical softmax}
\vspace{-0.5em}

\cite{chen2015strategies} benchmarked many different methods to train neural language models. Among the strategies they tried, they found that for very large vocabularies, the hierarchical softmax \citep{Morin+al-2005, Mikolov-et-al-NIPS2013} is the fastest and the best for a fixed budget of training time. Therefore we also compared the Z-loss to it. The hierarchical softmax modifies the original architecture by replacing the softmax by a heuristically defined hierarchical tree.

\vspace{-0.5em}
\section{The proposed Z-loss}
\vspace{-0.25em}

\begin{figure}[!ht]
\vspace{-0.5em}
\centering
\begin{minipage}[c]{4cm}
	\caption{Plot of the Z-loss $L_Z$ in function of $z_c = \frac{o_c - \mu}{\sigma}$ for $D=1000$, $a=0.1$ and $b=10$. The hyperparameter $a$ controls the softness of the $\softplus$. The dashed grey line represents the asymptote while $z_c$ tends to $-\infty$. $z_c$ is bounded between $-\sqrt{D-1}$ and $\sqrt{D-1}$.}
	\label{fig:softplus}
\end{minipage}
\begin{minipage}[c]{6cm}
	\includegraphics[width=6cm]{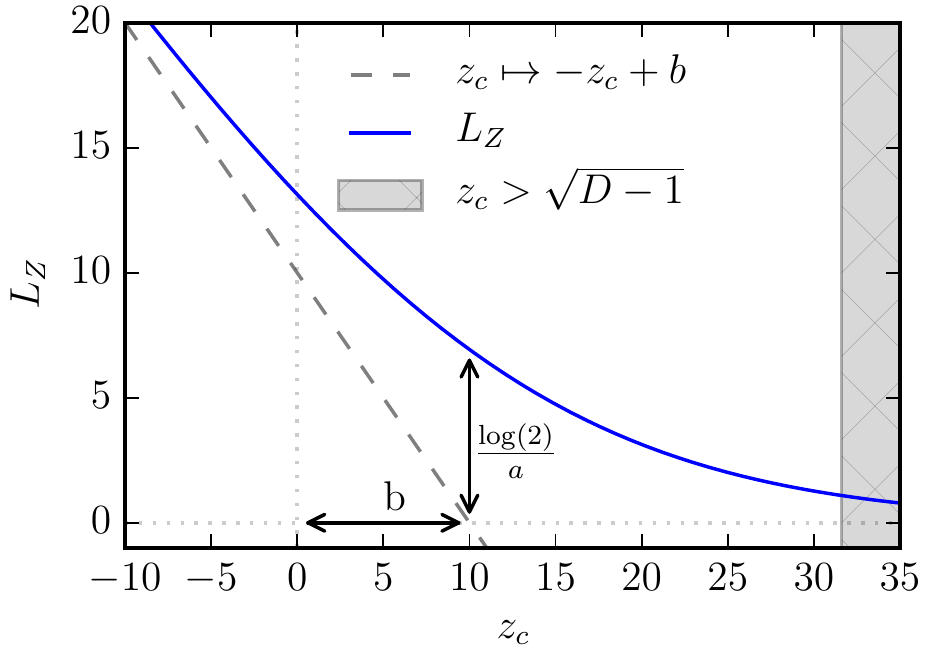}
\end{minipage}
\vspace{-1em}
\end{figure}

Let $\mu$ and $\sigma$ be the mean and the standard deviation of the pre-activations $\bm{o}$ of the current example: $\mu = \sfrac{\sum_{k=1}^{D} o_k }{D}$ and $\sigma^2 = \sfrac{\sum_{k=1}^{D} o_k^2 }{D} - \mu^2$. We define the \emph{Z-normalized} outputs $\bm{z}=[z_1,...,z_n]$ as $z_k = \frac{o_k - \mu}{\sigma}$, which we use to define the Z-loss as
\begin{align}
\label{eq:z_loss}
L_Z(\bm{o}, c) = L_Z(z_c) = \frac{1}{a} \softplus (a (b - z_c) ) = \frac{1}{a} \log \left[ 1 + \exp \left(a \left(b - \frac{o_c - \mu}{\sigma} \right) \right) \right],\end{align}
where $a$ and $b$ are two hyperparameters controlling the scaling and the position of the vector $z_c$. The Z-loss can be seen as a function of a single variable $z_c$ and is plotted in Figure~\ref{fig:softplus}. The Z-loss clearly belongs to the spherical family described in section~\ref{spherical_losses}. It can be decomposed into three successive operations: the normalization of $\bm{o}$ into $\bm{z}$ (which we call Z-normalization), the scaling/shift of $\bm{z}$ (controlled with $a$ and $b$) and the $\softplus$. Let us analyse these three stages successively.

\textbf{Z-normalization}: The normalization of $\bm{o}$ into $\bm{z}$, which we call Z-normalization, is essential in order to involve all the different output components $o_k$ in the final loss. Without it, the loss would only depend on $o_c$ and not on the other $o_k$, resulting in a null gradient with respect to the other $o_k$. Thus, thanks to the normalization, the pre-activations $\bm{o}$ compete against each other and there are three interlinked ways to increase $z_c$ (i.e. minimizing the loss): either increase $o_c$, or decrease $\mu$ or decrease $\sigma$. This behavior is similar to the log-softmax. Furthermore, this standardization makes the Z-loss invariant to both shifting and scaling of the outputs $\bm{o}$, whereas the log-softmax is only invariant to shifting. Note that the rank the classes is unaffected by global shifting and scaling of the pre-activations $o$, and so are any rank-based \emph{task losses} such as precision at $k$. Since the Z-loss is similarly invariant, while the log-softmax is sensitive to scale, this may make the Z-loss a better surrogate for rank-based \emph{task losses}. 

The gradient of the Z-loss with respect to $\bm{o}$ is simply the gradient of $z_c$ times the derivative of the softplus. The gradient of $z_c$ with respect to $\bm{o}$ can be written as
\begin{align*}
\frac{\partial z_c}{\partial o_{c}} = \frac{1}{D \sigma} \left( z_c^2 -D +1 \right) \ \text{ and } \ 
\frac{\partial z_c}{\partial o_{k}}_{k \neq c} = \frac{1}{D \sigma} \left( z_c z_k - 1 \right).
\end{align*}
The sum of the gradient components is zero, enforcing the pre-activations to compete against each other.
It equals zero when:
\begin{align*}
\frac{\partial z_c}{\partial o_{c}} = 0 \Leftrightarrow z_c^2 = D-1 \ \text{ and } \ 
\frac{\partial z_c}{\partial o_{k}}_{k \neq c} = 0 \Leftrightarrow z_k = \frac{1} {z_c}.
\end{align*}
Therefore $z_c$ is bounded between $-\sqrt{D-1}$ and $\sqrt{D-1}$. The gradient of the Z-loss with respect to $\bm{o}$ is simply the gradient of $z_c$ times the derivative of the softplus, which is $\sig$:
\begingroup\makeatletter\def\f@size{9}\check@mathfonts
\begin{align*}
\frac{\partial L_Z}{\partial o_{c}} = \frac{1}{D \sigma} \left( z_c^2 -D +1 \right) \sig (a (b - z_c)) \ \text{ and } \ 
\frac{\partial L_Z}{\partial o_{k}}_{k \neq c} = \frac{1}{D \sigma} \left( z_c z_k - 1 \right) \sig (a (b - z_c)),
\end{align*}
where $\sig$ denotes the logistic sigmoid function defined as $ \sig(x) = \frac{1}{1+\exp(-x)}$. Like $z_c$, the components still sum to one and the Z-loss reaches its minimum when $z^*_c=\sqrt{D-1}$ and $\forall k \neq c: z^*_k = -\frac{1}{\sqrt{D-1}}$, for which an infinite number of corresponding $\bm{o}$ vectors are possible (if $\bm{o}^*$ is solution, then for any $\alpha$ and $\beta$, $\alpha \bm{o}^* +\beta$ is also solution). Unlike the Z-loss, the log-softmax does not have such fixed points and, as a result, its minimization could potentially push $\bm{o}$ to extreme values.

Note that this Z-normalization is different from that used in batch normalization \citep{Ioffe+Szegedy-2015}. Ours applies across the dimensions of the output for each example, whereas batch normalization separately normalizes each output dimension across a minibatch.

\textbf{Scaling and shifting}: The normalized activations $\bm{z}$ are then scaled and shifted by the affine map $\bm{z} \mapsto a ( \bm{z} - b)$. These two hyperparameters are essential to allow the Z-score to better match the \emph{task loss}, which we are ultimately interested in. In particular, we will see later that the optimal values of these parameters significantly vary depending on the specific \emph{task loss} we aim to optimize. $a$ controls the softness of the $\softplus$, a large $a$ making the $\softplus$ closer to the rectifier function ($x \mapsto max(0,x)$). Note that the effect of changing $a$ and $b$ cannot be cancelled by correspondingly modifying the output layer weights $W$. This contrasts with the other classic loss functions, such as the log-softmax, for which the effect could be undone by reciprocal rescaling of $W$ as discussed further in Section~\ref{sec:extra_param}.

\textbf{Softplus}:
\label{softplus}
The $\softplus$ ensures that the derivative with respect to $z_c$ tends towards zero as $z_c$ grows. Without it, the derivative would always be $-1$, which would strongly push $z_c$ towards extreme values (still bounded by $\sqrt{D-1}$) and potentially employ unnecessary capacity of the network. We can also motivate the choice of using a $\softplus$ function by deriving the Z-loss from a multi-label classification perspective (non-mutually-exclusive classes). Let $\mathcal{Y}$ be the random variable representing the class of an example, it can take values between $1$ and $D$. Let us consider now the multi-label setup in which we aim to model each output $Y_k = \delta_{\mathcal{Y}=k}$ as a Bernoulli law whose parameter is given by a sigmoid $P(Y_k = 1) = \sig(o_k)$. Then, the probability of class $c$ can be written as the probability of $Y_c$ being one times the probabilities of the other $Y_k$ being zero: $P(\mathcal{Y}=c) = P(Y_c=1) \prod_{k\neq c} P(Y_k=0)$. Minimizing the negative log-likelihood of this model leads to the following cross-entropy-sigmoid loss:
\begin{align*}
L_{CE}(\bm{o}, c) = -\log(P(\mathcal{Y}=c)) = \softplus(-o_c) + \sum_{k \neq c} \softplus(o_k).
\end{align*}
If we only minimize the first term and ignore the others, the values of $\bm{o}$ would systematically decrease and the network would not learn. If instead we keep only the first term but apply the Z-normalization beforehand, we obtain the Z-loss, as defined in equation~\ref{eq:z_loss}. We claim that the Z-normalization compensates the approximation, as the ignored term is more likely to stay approximatively constant because it is now invariant to shift and scaling of $\bm{o}$. In our experiments, we will evaluate the $L_{CE}$ along the Z-loss.

%



\textbf{Generaliszation: Z-normalization before any classic loss functions}:
\label{sec:extra_param}

The Z-normalization could potentially be applied to any other classic loss functions (the resulting loss functions would always be scale and shift invariant). Therefore, we also compared the Z-loss to the Z-normalized version of the log-softmax $L_S$. The shifting parameter $b$ is useless as the softmax is shift-invariant. We denote $L_{S-Z}$ the corresponding Z-normalized loss function:
\begin{align*}
L_{S-Z}(\bm{o}, c) = -\frac{1}{a} \log(\softmax_c(a \bm{z})).
\end{align*}
Note that this is different from simply scaling the output activations $\bm{o}$ with $a$: $L(\bm{o}, c) = -\frac{1}{a} \log(\softmax_c(a \bm{o}))$. In that latter case, contrary to $L_{S-Z}$, the effect of $a$ could be undone by reciprocal rescaling of $W$.

%

\section{Experiments}

\begin{figure*}[!ht]
	\includegraphics[width=\textwidth]{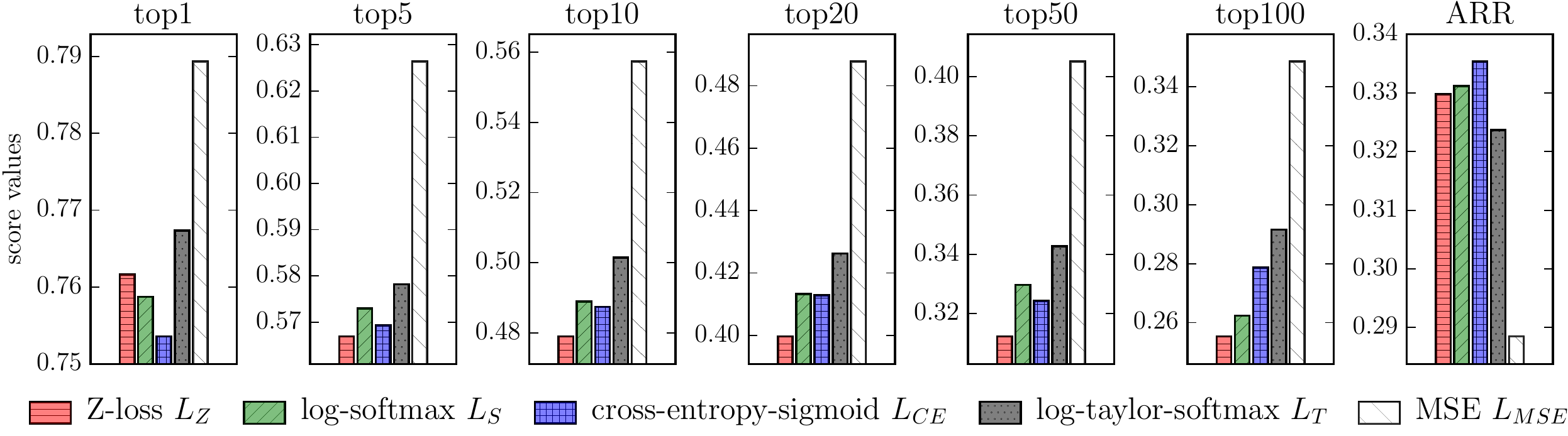}
	\vspace{-1.5em}
	\caption{Top-k error rates and Mean Reciprocal Rank (MRR, equivalent to the Mean Average Precision) obtained by our best models for each loss function on the Penn Tree Bank language modeling task. The Mean-Square-Error (MSE) has the worst performance, followed by the Taylor-Softmax. The cross-entropy-sigmoid has the lowest top-1 error rate and surprisingly outperforms the log-softmax. The Z-loss has the lowest top-\{5,10,20,50,100\} error rates (the hyperparameters $a$ and $b$ were tuned individually for each top-k).}
	\label{fig:penn_tree_scores}
\end{figure*}

\cite{brebisson2016spherical} already conducted experiments with several spherical losses (the Taylor/Spherical softmax and the Mean Squared Error) and showed that, while they work well on problems with few classes, they are outperformed by the log-softmax on problems with a large number of output classes. Therefore we focused our experiments on those problems and in particular on word-level language modeling tasks for which large datasets are publicly available. The task of word-language modeling consists in predicting the next word following a sequence of consecutive words called a $n$-gram, where $n$ is the length of the sequence. For example, "A man eats an apple" is a 5-gram and "A man eats an" can be used as an input sequence context to predict the target word "apple". Neural language models~\citep{nnlm:2001:nips} tackles this classification task with a neural network, whose number of outputs is the size of the vocabulary.

As the Z-loss does not produce probabilities, we cannot compute likelihood or perplexity scores comparable to those naturally computed with the log-softmax model. Therefore we instead evaluated our different loss functions on the following scores (which are often considered as the ultimate \emph{task losses}): top-\{1,5,10,20,50,100\} error rates and the mean reciprocal rank (equivalent to the mean average precision in the context of multi-class classification), defined below. Let $r_c$ be the rank of the pre-activation $o_c$ among $\bm{o}$, it can take values in $[1,...,D]$. If $r_c=1$, the point is well-classified.

\textbf{Top-k error rate}: The top-k error rate is defined as the mean of the boolean random variable defined as $r_c \le k$. It measures how often the target is among the $k$ highest predictions of the network.

\textbf{Mean Reciprocal Rank (MRR)}: It is defined as the mean of $\frac{1}{r_c}$. A perfect classification would lead to $r_c = 1$ for all examples and thus an MRR of $1$. The MRR is identical to the Mean Average Precision in the context of classification. These are popular score measures for ranking in the field of information retrieval.

\vspace{-1em}
\subsection{Penn Tree bank}
\vspace{-0.5em}


We first trained word-level language models on the classic Penn tree bank~\citep{marcus1993building}, which is a corpus split into a training, validation and testing set of 929k words, a validation set of 73k words, and a test set of 82k words. The vocabulary has 10k words. We trained typical feed-forward neural language models  with vanilla stochastic gradient descent on mini-batches of size 250 using an input context of 6 words. For each loss function, we tuned the embedding size, the number of hidden layers, the number of neurons per layer, the learning rate and the hyperparameters $a$ and $b$ for the Z-loss. Figure~\ref{fig:penn_tree_scores} reports the final test scores obtained by the best models for each loss and each evaluation metric. As can be seen, the Z-loss significantly outperforms the other considered losses with respect to the top-\{5,10,20,50,100\} error rates.

To measure to which extent the hyperparameters $a$ and $b$ control how the Z-loss matches the \emph{task losses}, we trained several times the same architecture for different values of $a$. The results are reported in Figure~\ref{fig:evolution_a}. Figure~\ref{fig:pt_training_profiles} shows the training curves of our best Z-score models for the top-\{1,5,10,50\} error rates respectively. We can see that the hyperparameters $a$ and $b$ drastically modify the training dynamics and they are thus extremely important to fit the particular evaluation metric of interest.

\begin{figure}[!htb]
\vspace{-0.5em}
\begin{minipage}[c]{0.4\textwidth}
	\caption{top-\{1,5,10\} error rates for Z-loss models trained on the Penn Tree Bank dataset which differ only in the value of their hyperparameter $a$. More precisely, for each value of $a$, a separate model has been trained from scratch. $b=28$ for all models. The three curves have very different shapes with different minima, showing that $a$ (and $b$ to a lesser extent) gives to the \emph{surrogate} Z-loss the flexibility to better fit the \emph{task loss}.}
	\label{fig:evolution_a}
\end{minipage}\hfill
\begin{minipage}[c]{0.59\textwidth}
	\hspace{-0em}\includegraphics[width=\textwidth]{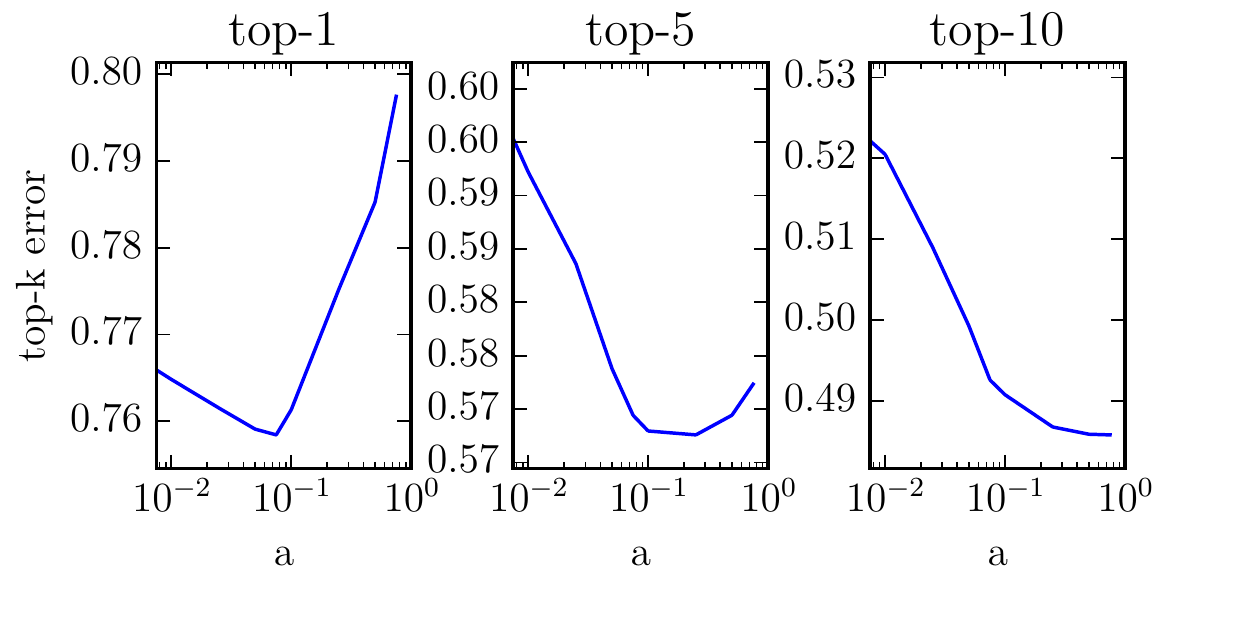}
\end{minipage}
\vspace{-0.4em}

\end{figure}

\begin{figure}[!htb]
	\includegraphics[width=\textwidth]{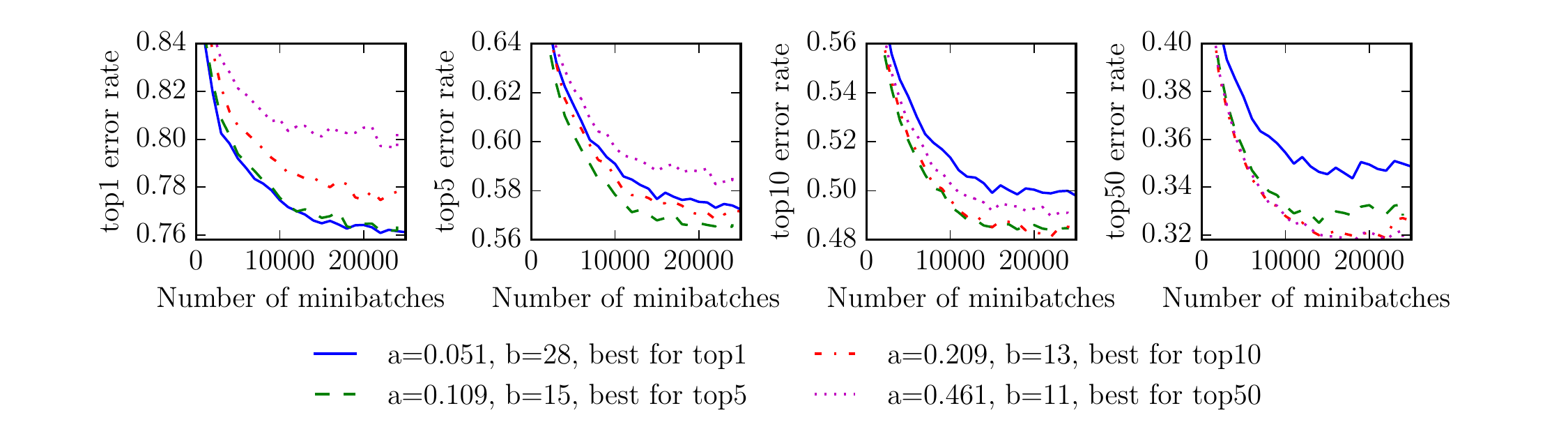}
	\vspace{-1.5em}
	\caption{Evolution of the validation top-\{1,5,10,50\} error rates during training on the Penn Tree Bank dataset for four Z-loss language models with different combinations of hyperparameters $a$ and $b$. Each of the four combinations has been chosen to minimize a particular top-k error rate. For example, the dashed green curve corresponds to the best model obtained with respect to the top-5 error rate. In particular we can see that the best top-1 model is the worst top-50 model and vice versa. The very high variations between top-k plots show how the hyperparameters $a$ and $b$ allow to better match the \emph{task loss}. In contrast, the classic log-softmax lack these flexibility hyperparameters.}
	\label{fig:pt_training_profiles}
\end{figure}

\begin{figure}[!htb]
\begin{minipage}[c]{0.4\textwidth}
	\caption{Comparison of top-\{1,10,50\} test error rates obtained by our best models for the Z-loss and loss functions with hyperparameters on the Penn Tree Bank language modeling task. The hyperparameters added to the log-softmax and cross-entropy do not seem to have an effect as important as for the Z-loss but still improve slightly the final scores. For the Z-loss, the Z-normalization is crucial and removing it would prevent any meaningful learning.}
	\label{fig:penn_tree_scores_normalized}
\end{minipage}\hfill
\begin{minipage}[c]{0.55\textwidth}
	\includegraphics[width=\textwidth]{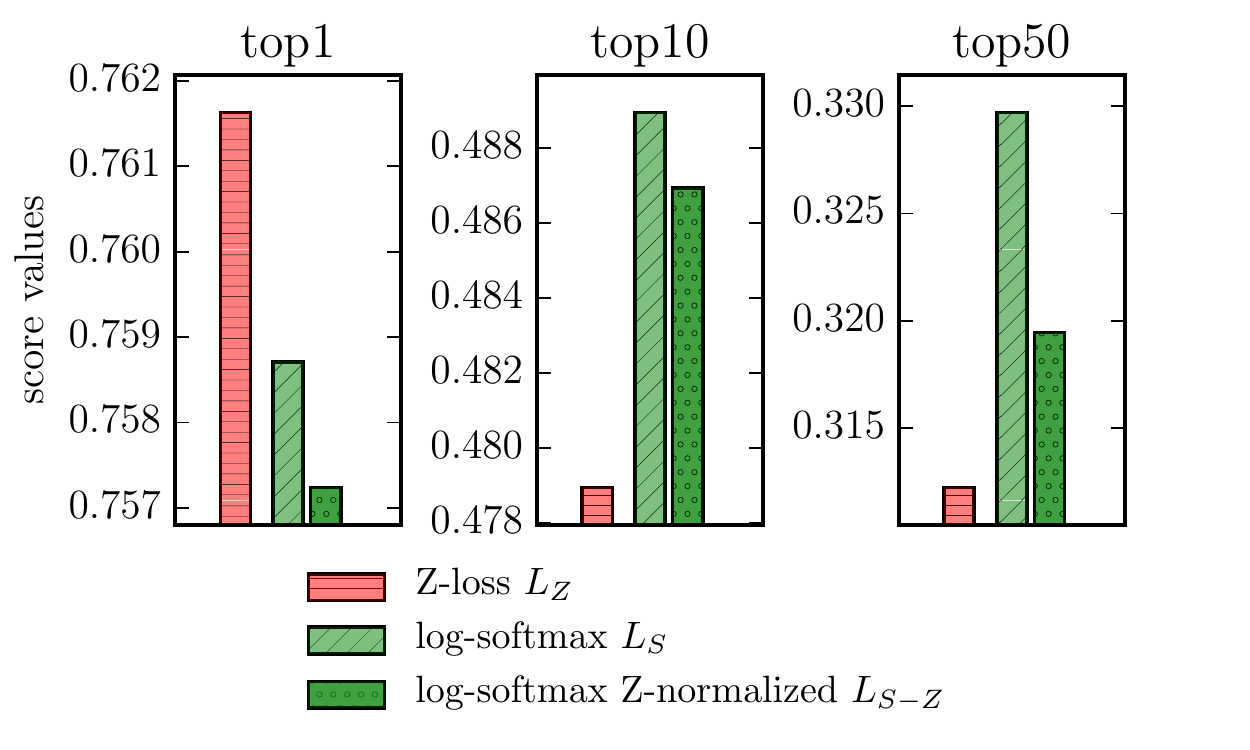}
\end{minipage}

\vspace{-1em}
\end{figure}

Figure~\ref{fig:penn_tree_scores_normalized} reports the test scores obtained by our best Z-normalized versions of the log-softmax. As previously explained in section~\ref{sec:extra_param}, the Z-normalization enables adding scaling hyperparameters $a$ and $b$, which also help the log-softmax to better match the the top-k evaluation metrics but not as much as the Z-loss.

\vspace{-1em}
\subsection{One Billion Word}
\label{sec:billion_word}
\vspace{-0.5em}

We also trained word-level neural language models on the One Billion Word dataset \citep{ChelbaMSGBKR14}, a considerably larger dataset than Penn Tree Bank. It is composed of 0.8 billion words belonging to a vocabulary of 793 471 words. Given the size of the dataset, we were not able to extensively tune the architecture of our models, nor their hyperparameters. Therefore, we first compared all the loss functions on a fixed architecture, which is almost identical to that of~\cite{chen2015strategies}: 10-grams concatenated embeddings representing a layer of 512*10=5120 neurons and three hidden layers of sizes 2048, 2048 and 512. We will refer to this architecture as \texttt{net1}. From our experiments and those of~\cite{chen2015strategies}, we expect that more than 40 days would be required to train \texttt{net1} with a naive log-softmax layer until convergence (on a high-end Titan X GPU). Among the workarounds that \cite{chen2015strategies} benchmarked, they showed that the hierarchical softmax is the fastest and best method for a fixed budget of time. Therefore, we only compared the Z-loss to the hierarchical softmax (a two-layer hierarchical softmax, which is most efficient in practice due to the cost of memory accesses). The architecture \texttt{net1} being fixed, we only tuned the initial learning rate for each loss function and periodically decreased it when the validation score stoped improving. Table~\ref{tab:billion_timings} and~\ref{tab:billion_scores} report the timings and convergence scores reached by the three loss functions with architecture \texttt{net1}. Although the hierarchical softmax yields slightly better top-k performance, the Z-loss model is more than 4 times faster to converge. This allows to train bigger Z-loss models in the same amount of time as the hierarchical softmax, and thus we trained a bigger Z-loss model with an architecture \texttt{net2} of size [1024*10= 10240, 4096, 4096, 1024, 793471] in less than the 4.08 days required by the hierarchical softmax with architecture \texttt{net1} to converge. As seen in table~\ref{tab:billion_scores}, this new model has a better top-1 error rate than the hierarchical softmax after only 3.14 days. It is very likely that another set of hyperparameters (a, b) would yield lower top-20 error rates as well.

\begin{table*}[!ht]
\vspace{-0.5em}
\caption{Training timings to process 1 epoch over the training data of the One Billion Word dataset (around 150 millions n-grams) for the different loss functions with the architecture \texttt{net1}, i.e. a feedforward network composed of 5 layers of sizes [5120, 2048, 2048, 512, 793471], with a batch size of 200. The GPU used is an Nvidia Titan X while the CPU is an Intel i7-5930K CPU @ 3.50GHz. We give the timings for both the whole model and the output layer only. We only timed a few thousands minibatches and extrapolated the timings for the whole epoch.
}
\label{tab:billion_timings}
\begin{center}
\begin{tabular}{| l || c | c | c | c |} 
 \cline{2-5}
 \multicolumn{1}{c||}{} 
 & \multicolumn{2}{|c|}{\centering Timings CPU} 
 & \multicolumn{2}{|c|}{\centering Timings GPU} \\ [0.5ex] 
 \hline
 \multicolumn{1}{|l||}{\centering Loss function} 
 & \multicolumn{1}{|p{2cm}|}{\centering whole model} 
 & \multicolumn{1}{|p{2cm}|}{\centering output only} 
  & \multicolumn{1}{|p{2cm}|}{\centering whole model}
 & \multicolumn{1}{|p{2cm}|}{\centering output only} \\ [0.5ex] 
 \hline\hline
 softmax & 78.5 days & 69.7 days &  4.56 days & 4.44 days \\ 
 \hline
 H-softmax & / & / & 12.23 h & 10.88 h \\ 
 \hline
 Z-loss & 7.50 days & 8.68 h & 2.81 h & 1.24 h \\ 
 \hline 
\end{tabular}
\end{center}
\vspace{-1em}
\end{table*}

\begin{table}[!ht]
\vspace{-0.5em}
\caption{
Final test top-1 and top-20 error rates on the One Billion Word language modeling task. The "Constant" line corresponds to a constant classifier predicting the frequencies of the words. The hierarchical softmax reaches a final perplexity of 80 (competitive with~\cite{chen2015strategies}). The hyperparameters $a$ and $b$ of the Z-loss model with both architectures \texttt{net2} and \texttt{net1} have been tuned to maximize the top-1 error rate. The GPU used is an Nvidia Titan X.}
\label{tab:billion_scores}
\begin{center}
\begin{tabular}{| l || c | c | c | c |} 
 \hline 
 \multicolumn{1}{|p{2cm}||}{Loss function}
 &\multicolumn{1}{|p{1.7cm}|}{\centering Architecture} 
 &\multicolumn{1}{|p{2.4cm}|}{\centering Top-1 error rate } 
 &\multicolumn{1}{|p{2.4cm}|}{\centering Top-20 error rate }
  &\multicolumn{1}{|p{2.6cm}|}{\centering Total training time}
  \\ [0.5ex] 
 \hline\hline
 Constant & / & 95.44 \% & 65.58 \% & / \\
 \hline  \hline
 Softmax & \texttt{net1} & / & / & about 40 days \\
 \hline 
 H-softmax & \texttt{net1} & 71.0 \% & \textbf{35.73} \% & 4.08 days \\
 \hline
 Z-loss & \texttt{net1} & 72.13 \% & 36.43 \% & \textbf{0.97 days} \\
 \hline \hline 
 Z-loss & \texttt{net2} & \textbf{70.77} \% & 38.29 \% & 3.14 days \\
 \hline 
\end{tabular}
\end{center}
\vspace{-0.3em}
\begin{addmargin}[3em]{2em}
\texttt{net1}: 5 layers of sizes [10*512, 2048, 2048, 512, 793471], batch size of 200,

\texttt{net2}: 5 layers of sizes [10*1024, 4096, 4096, 1024, 793471], batch size of 1000.
\end{addmargin}
\vspace{-0.5em}
\end{table}

\section{Discussion}

The cross-entropy sigmoid outperforms the log-softmax in our experiments on the Penn Tree Bank dataset with respect to the top-\{1,5,10,20,50\} error rates. This is surprising because the cross-entropy sigmoid models a multi-label distribution rather than a multi-class one. This might explain why the Z-loss, which can be seen as an approximation of the cross-entropy sigmoid (see Section~\ref{softplus}), performs so well: it is slightly worse than the log-softmax for the top-1 error but outperforms both the softmax and the cross-entropy sigmoid for the other top-k. It very significantly outperforms the other investigated spherical loss functions, namely the Taylor softmax and the Mean Square Error.

Our results show that the two hyper-parameters $a$ and $b$ of the Z-loss are essential and allow it to fit certain evaluation metrics (such as top-k scores) more accurately than the log-softmax. We saw that we can also add hyperparameters to any traditional loss function by applying the Z-normalization beforehand. In particular these hyperparameters slightly improve the performance of the log-softmax even though their effect is not as important as with the Z-loss (Figure~\ref{fig:penn_tree_scores_normalized}). In practice, the hyperparameters of the Z-loss are simple to tune, we found that running the search on the first iterations is sufficient. For the top-k error rates, the hyperparameter $a$ is more important than $b$: the higher it is, the better the top-$k$ scores with a high $k$ and vice versa.

On the One Billion Word language modeling task, the Z-loss models train considerably faster than the hierarchical softmax (a 4x speedup for the identical architecture \texttt{net1}) but is slightly worse with respect to the final top-k scores. Thanks to the speed of the Z-loss, we were able to train a significantly larger architecture (\texttt{net2}) faster than the hierarchical softmax on a smaller architecture (\texttt{net1}) and obtain slightly better top-1 error rate. The Z-loss top-20 score is not as good because the hyperparameters $a$ and $b$ were tune for the top-1.
\section{Conclusion}

We introduced a new loss function, the Z-loss, which aims to address two potential limitations of the naive log-softmax: the speed when the problem has a large amount of output classes and the discrepancy with the \emph{task loss} that we are ultimately interested in. Contrary to the log-softmax, the Z-loss has the desirable property of belonging to the spherical family, which allows to train the output layer efficiently, independently from the number of classes\footnote{ The source code of our efficient Z-loss implementation is available online: https://github.com/pascal20100/factored\_output\_layer.}. On the One Billion Word dataset with 800K classes, for a fixed standard network architecture, training a Z-loss model is about 40 times faster than the naive log-softmax version and more than 4 times faster than the hierarchical softmax. For a fixed budget of around 4 days, we were able to train a better Z-loss model than the hierarchical softmax with respect to the top-1 error rate. Complexity-wise, if $D$ is the number of classes, the computations of the hierarchical softmax scale in $log(D)$ in theory (in practice $\sqrt{D}$ for a memory-efficient 2-layer hierarchical softmax implementation), while those of the Z-loss are independent from the output size $D$. This suggests that the Z-loss would be better suited for datasets with even more classes, on which the hierarchical softmax would be too slow.

In addition to the huge speedups, the Z-loss also addresses the problem of the discrepancy between the \emph{task loss} and the \emph{surrogate loss}. Thanks to a shift and scale invariant Z-normalization, the Z-loss benefits from two hyperparameters that can adjust, to some extent, how well the \emph{surrogate} Z-loss matches the \emph{task loss}. We showed experimentally that these hyperparameters can drastically improve the resulting \emph{task loss} values, making them very desirable. On the Penn Tree Bank, our Z-loss models yield significantly lower top-\{5,10,20,50,100\} error rates than the log-softmax. Further research will focus on updating these hyperparameters automatically \emph{during training} to ensure that the loss function dynamically matches the \emph{task loss} as close as possible. Beyond the Z-loss, the Z-normalization is interesting on its own and can be applied to any classic loss functions, such as the log-softmax, allowing to add hyperparameters to any loss function and potentially mitigating the discrepancy with the \emph{task loss}. Further research should investigate generalizations of the Z-normalization in a more general framework than the Z-loss.

\newpage

\bibliography{strings,strings-shorter,ml,zoulou,aigaion-shorter}
\bibliographystyle{unsrtnat_initials_no_url_editors}
\end{document}